\documentclass[letterpaper]{article} % DO NOT CHANGE THIS
\usepackage{aaai2026}  % DO NOT CHANGE THIS
\usepackage{times}  % DO NOT CHANGE THIS
\usepackage{helvet}  % DO NOT CHANGE THIS
\usepackage{courier}  % DO NOT CHANGE THIS
\usepackage[hyphens]{url}  % DO NOT CHANGE THIS
\usepackage{graphicx} % DO NOT CHANGE THIS
\usepackage{natbib}  % DO NOT CHANGE THIS AND DO NOT ADD ANY OPTIONS TO IT
\usepackage{caption} % DO NOT CHANGE THIS AND DO NOT ADD ANY OPTIONS TO IT
\usepackage[ruled,linesnumbered]{algorithm2e}
\usepackage{appendix}
\usepackage{booktabs}       % professional-quality tables
\usepackage{multirow}
\usepackage{multicol}
\usepackage{tikz}
\usepackage{amsmath, amssymb, amsfonts}
\usetikzlibrary{trees, positioning}
\title{PA-FAS: Towards Interpretable and Generalizable Multimodal Face Anti-Spoofing via Path-Augmented Reinforcement Learning}
\author{
    Yingjie Ma\equalcontrib\textsuperscript{\rm 1,2},
    Xun Lin\equalcontrib\textsuperscript{\rm 2},
    Yong Xu\textsuperscript{\rm 5},
    Weicheng Xie\textsuperscript{\rm 1,3},
    Zitong Yu\textsuperscript{\rm 2,3,4}\thanks{Corresponding author}
}
\affiliations{
    \textsuperscript{\rm 1}College of Computer Science and Software Engineering, Shenzhen University\\
    \textsuperscript{\rm 2}School of Computing and Information Technology, Great Bay University\\
    \textsuperscript{\rm 3}Guangdong Provincial Key Laboratory of Intelligent Information Processing \& Shenzhen Key Laboratory of Media Security, Shenzhen University\\
    \textsuperscript{\rm 4}Dongguan Key Laboratory for Intelligence and Information Technology\\
    \textsuperscript{\rm 5}Harbin Institute of Technology, Shenzhen
}

\usepackage{bibentry}
\begin{document}

\maketitle

\begin{abstract}

% 面部防伪（Face Anti-Spoofing, FAS）是保障人脸识别系统在复杂环境中安全性和可靠性的重要环节。近年来，FAS在多模态融合、跨域泛化以及可解释性等方向取得了显著进展，但这三方面仍缺乏统一的建模框架，限制了面部防伪技术的可信性与实际部署能力。随着大模型和强化学习技术的发展，基于策略优化的训练范式为实现多模态、泛化与可解释性的协同建模提供了新的可能。
% 然而，多模态推理相较于单模态推理，增加了对不同模态的准确特征描述和模态互证等复杂逻辑，这不仅显著提升了推理复杂度，也大幅增加了标注难度。由于现有的多模态FAS数据集缺乏高质量标注信息，直接应用强化学习策略难以奏效，从而限制了有效的多模态推理。
% 为探究监督微调和强化学习（SFT+RL）范式在多模态FAS推理任务中失败的具体原因，我们分析并发现了两个关键问题：一是多模态推理路径有限，不仅无法充分利用多模态信息，还限制了SFT后模型的探索空间，影响后续RL的充分探索；二是单一任务类型与丰富的多模态推理路径易导致推理困惑，模型可能形成仅基于输入图像直接输出答案的捷径，使推理无效。这些问题进一步加剧了多模态推理的复杂性，导致强化学习策略难以有效应用。
% 为此，我们提出一种面向面部防伪（FAS）任务的推理路径增强方法。该方法基于有限标注数据构建高质量扩展推理序列，丰富训练数据中的有效推理路径，缓解了探索空间不足的问题。同时，在SFT阶段引入答案随机打乱机制，通过随机调整部分训练样本的答案标签，迫使模型在训练过程中不能仅依赖输入图像的直接特征，而是需要通过多模态信息的综合分析来推导正确答案，从而专注于推理路径的扩展，避免形成捷径。
% 我们提出的方法有效地改善了上述提到的问题与挑战，显著提升了多模态推理的准确性和泛化能力，成功整合了多模态融合、跨域泛化和可解释性三个方面，对面部防伪领域具有重要意义。
%Face Anti-Spoofing (FAS) is a critical component for ensuring the security and reliability of face recognition systems in complex environments. 
%In this paper, we find two key issues of supervised fine-tuning combined with reinforcement learning (SFT+RL) paradigm multimodal FAS inference: 1) limited reasoning paths after SFT constrain the model’s exploration space, reducing the effectiveness of subsequent RL; and 2) the mismatch between single-task supervision and diverse reasoning paths leads to reasoning confusion, where models may exploit shortcuts by directly mapping input images to answers, bypassing the intended reasoning process. 
%However, the lack of a unified modeling framework across these three aspects limits the trustworthiness and practical deployment of FAS systems. 
Face anti-spoofing (FAS) has recently advanced in multimodal fusion, cross-domain generalization, and interpretability. With large language models and reinforcement learning (RL), strategy-based training offers new opportunities to jointly model these aspects. However, multimodal reasoning is more complex than unimodal reasoning, requiring accurate feature representation and cross-modal verification while facing scarce, high-quality annotations, which makes direct application of RL sub-optimal. We identify two key limitations of supervised fine-tuning plus RL (SFT+RL) for multimodal FAS: (1) limited multimodal reasoning paths restrict the use of complementary modalities and shrink the exploration space after SFT, weakening the effect of RL; and (2) mismatched single-task supervision versus diverse reasoning paths causes reasoning confusion, where models may exploit shortcuts by mapping images directly to answers and ignoring the intended reasoning. To address this, we propose PA-FAS, which enhances reasoning paths by constructing high-quality extended reasoning sequences from limited annotations, enriching paths and relaxing exploration constraints. We further introduce an answer-shuffling mechanism during SFT to force comprehensive multimodal analysis instead of using superficial cues, thereby encouraging deeper reasoning and mitigating shortcut learning. PA-FAS significantly improves multimodal reasoning accuracy and cross-domain generalization, and better unifies multimodal fusion, generalization, and interpretability for trustworthy FAS.

\end{abstract}

\begin{figure}[t]
\vspace{-1.5em}
\centering
\includegraphics[width=0.49\textwidth]{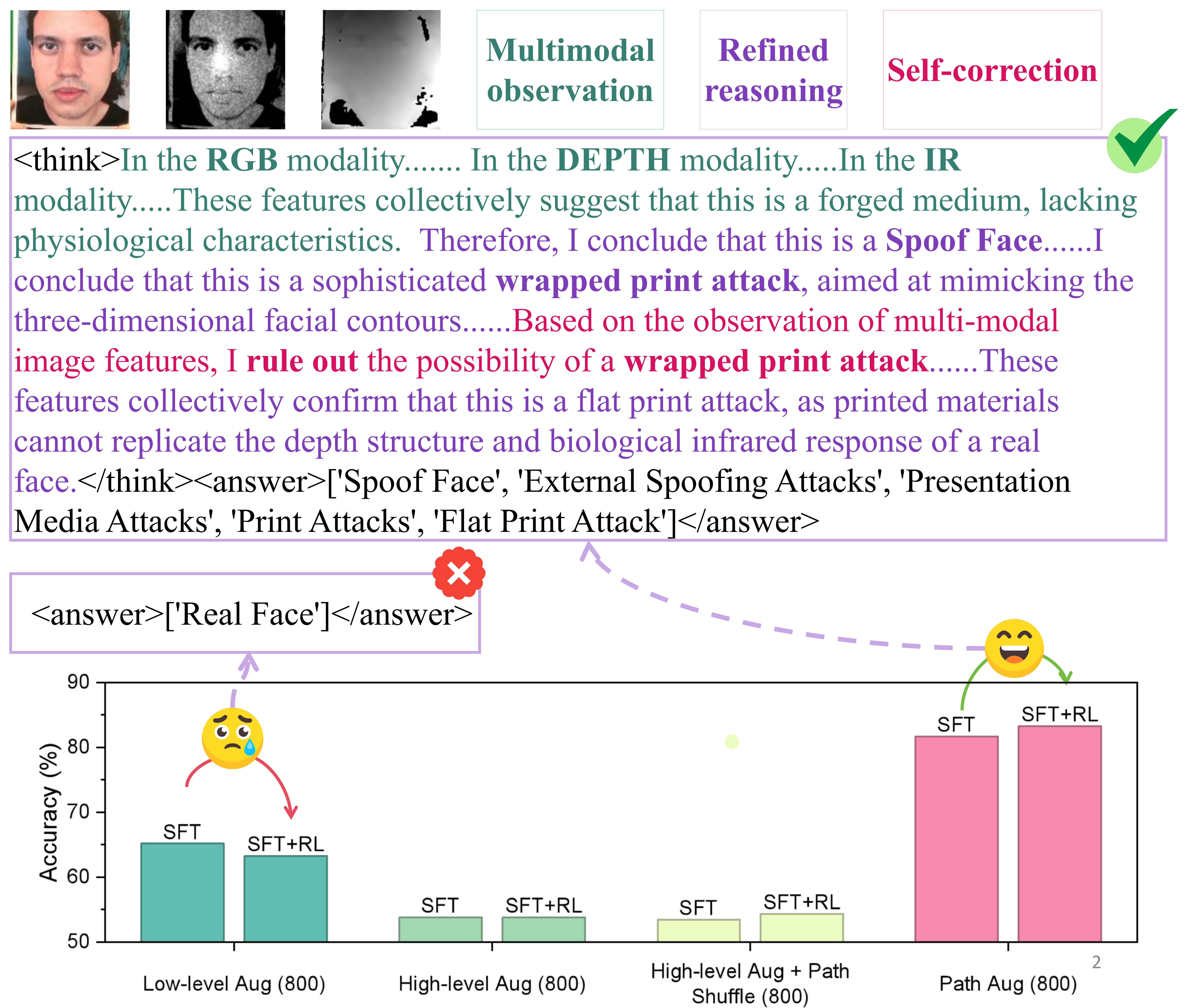}
\vspace{-1.5em}
%不同数据集在 SFT 和 SFT+RL 方法下的准确率示意图。在固定 800 样本量的条件下，单一推理路径的数据集在经过 SFT 训练后，于后续的 RL 阶段难以进一步提升准确率，甚至可能出现性能下降的情况。而具有丰富推理路径的数据集则在 SFT 和 SFT+RL 方法中均展现出显著的性能优势，能够实现更高的准确率。
\caption{\small{Accuracy of SFT and SFT+RL methods on different augmented datasets. With a fixed data size of 800, datasets with a single reasoning path fail to achieve higher accuracy in the subsequent RL stage after SFT training, and may even experience a decline in performance. In contrast, datasets with diverse reasoning paths demonstrate significantly better performance, achieving higher accuracy under both SFT and SFT+RL methods.}}
\label{fig:fig1}
\vspace{-1.3em}
\end{figure}

\vspace{-1.0em}
\section{Introduction}
% 人脸识别（Face Recognition, FR）系统广泛应用于支付、身份验证、安防监控等场景，但由于其高度依赖视觉信息，易受到如打印照片、视频回放、3D面具等呈现攻击（Presentation Attacks, PAs）的威胁，带来了严重的安全隐患。为提升系统鲁棒性，面部防伪（Face Anti-Spoofing, FAS）技术应运而生，旨在识别真实人脸与伪造攻击之间的差异。传统FAS方法\cite{yu2022deep}主要基于RGB图像，在可获取信息有限的条件下构建分类模型，难以应对复杂多样的攻击手段。为增强判别能力，近年来多模态FAS方法\cite{yu2023flexible}结合RGB、深度与红外信息，引入空间几何结构与热特征，显著提升了模型的识别准确性与鲁棒性。然而，复杂模态之间的融合与理解也带来了新的挑战，尤其是在安全敏感应用中，模型的泛化能力与可解释性显得尤为关键。
Face recognition (FR) systems have been widely adopted in scenarios such as payment authentication, identity verification, and surveillance. However, due to their heavy reliance on visual information, they are highly vulnerable to Presentation Attacks (PAs), including printed photos, replayed videos, and 3D masks, posing significant security risks. To enhance system robustness, Face Anti-Spoofing (FAS) techniques have emerged, aiming to distinguish between genuine and spoofed facial presentations. Traditional FAS methods~\cite{yu2022deep}, primarily based on RGB modality, struggle to cope with increasingly diverse and sophisticated attack modalities due to limited information. Recently, multimodal FAS approaches~\cite{yu2023flexible} incorporating DEPTH and infrared modalities alongside RGB have demonstrated significant improvements in both accuracy and robustness. However, these advances bring new challenges in integrating and interpreting heterogeneous modalities in real-world FAS applications where generalization ability and interpretability are crucial.
%introducing spatial-geometric and thermal cues,

% 当前FAS研究面临空白：域泛化能力（Domain Generalization, DG）不足、多模态（Multimodal, MM）方法缺乏解释、以及MLLM（Multimodal Large Language Models）方法难以扩展至泛化场景。在DG方面，尽管已有方法在提升模型跨场景、跨攻击类型的能力方面取得进展\cite{iadg2023,safas2023}，但它们多基于单模态设计，难以适用于多模态情境，且推理过程缺乏可解释性支撑。而多模态FAS方法\cite{fmvit2023}虽然性能优越，却缺少对深度与红外模态中欺骗线索的有效解释机制，模态间复杂的推理逻辑（如模态互证、模态冲突）也尚未被系统建模，影响了其可信度和实际部署能力。与此同时，MLLM近年来在单模态FAS中展现出较强的语言解释能力\cite{facecot2025,shield2025}，但现有工作几乎忽视了泛化能力的提升，同时尚未触及多模态线索的融合与推理，使得其难以直接扩展到现实中复杂的多模态防伪任务中。虽然近期有工作分别在尝试统合域泛化与多模态\cite{MMDG2024,mmdgplus2025,dadm2025,mmda2025}和统合可解释性和域泛化\cite{zhang2025interpretable}FAS任务并取得了优异的性能，且有些工作在MLLM中引入多模态数据集\cite{shield2025}来结合多模态与可解释性，但是在同时统合域泛化、多模态、可解释性上的研究仍然存在空白。
%Moreover, complex inter-modal reasoning logic (e.g., modality verification and conflicts) remains underexplored, limiting trustworthiness and real-world deployment. 
FAS research faces research gaps: 1) insufficient domain generalization (DG); 2) lack of interpretability in multimodal approaches; and 3) limited extensibility of Multimodal Large Language Models (MLLMs) to domain generalization scenarios. While existing methods have made progress in improving model robustness across domains and attack types~\cite{iadg2023,safas2023}, they are mostly developed for unimodal settings and offer limited insight into the model’s decision process. On the other hand, although recent multimodal methods~\cite{fmvit2023} show superior performance, they lack explicit interpretability mechanisms for identifying spoofing cues in DEPTH and infrared modalities. Meanwhile, recent MLLM-based FAS approaches~\cite{facecot2025,shield2025} have demonstrated strong language-level reasoning capabilities in unimodal scenarios. However, they neglect the generalization issue and fail to address cross-modal cue integration and reasoning for real-world multimodal spoofing detection. Although some recent works have attempted to unify domain generalization and multimodal FAS~\cite{MMDG2024,mmdgplus2025,dadm2025,mmda2025}, and others have explored the integration of interpretability and domain generalization in FAS tasks~\cite{zhang2025interpretable}, as well as leveraging multimodal datasets in MLLMs~\cite{shield2025} to bridge multimodality and interpretability. However, a comprehensive solution that simultaneously addresses domain generalization, multimodal fusion, and interpretability remains unexplored.

% 为增强大模型的推理能力，一些最新研究开始尝试引入链式思维（Chain-of-Thought, CoT）推理与强化学习（Reinforcement Learning, RL）范式，以增强模型的逻辑推理与跨域适应能力。尤其是小组相对策略优化（Group Relative Policy Optimization, GRPO）方法\cite{grpo2024}，通过采用基于规则的群体优势策略，避免了对昂贵神经奖励模型的依赖，在多个任务上展现出出色的泛化性能和较低训练开销，为构建具备可解释性域泛化能力的FAS系统提供了新路径。同时，SFT记忆机制与RL泛化策略的结合\cite{chu2025sft}为分阶段训练提供理论支持：通过在监督精调阶段（SFT）建立稳定记忆，再借助RL阶段进行策略探索与自我提升，有望打破模型过拟合训练数据的局限。
% 然而，在多模态面部防伪（FAS）推理任务中，SFT+RL范式的应用面临诸多挑战。这些挑战首先源于多模态推理本身的高复杂度：相比单模态任务，多模态FAS需要准确建模RGB、深度、红外等不同模态的特征，并实现模态互证、冲突解决和模态辅助等复杂推理逻辑，显著增加了模型训练和推理的难度。也正由于这一内在复杂性，高质量、多粒度、多样化的标注数据极难获取，现有数据集普遍仅提供单一模态输入与简单二元分类标签，缺乏对多模态关系的精细表达。如图\ref{fig:fig1}所示，在这种数据条件下，模型在SFT阶段容易陷入僵化学习与过度自信，无法真正掌握多模态之间的逻辑联系；而在RL阶段，由于监督信号单一且探索空间受限，模型难以获得有效反馈，策略学习变得低效甚至失效。最终，这些问题共同限制了SFT+RL训练范式在多模态FAS场景中的泛化能力与可解释性，甚至可能导致其性能不如单纯的SFT方法。
%This framework has the potential to overcome overfitting issues, but
% multimodal FAS reasoning tasks face critical challenges primarily due to the inherent complexity of multimodal inference itself. 
Recent studies have explored enhancing LLM reasoning capabilities via Chain-of-Thought (CoT) and Reinforcement Learning (RL) paradigms to improve logical reasoning and domain adaptability. Notably, the Group Relative Policy Optimization (GRPO) algorithm~\cite{grpo2024} introduces a rule-based group advantage strategy that avoids the need for expensive neural reward models, achieving impressive generalization with low training costs. This offers a new pathway toward constructing FAS systems with interpretable DG capabilities. Furthermore, the integration of Supervised Fine-Tuning (SFT) memory mechanisms with RL-based generalization strategies~\cite{chu2025sft} provides theoretical grounding for staged training: building stable knowledge during the SFT phase, followed by policy exploration and self-improvement in the RL phase. However, unlike unimodal settings, multimodal FAS requires accurate representation of RGB, DEPTH, and infrared modalities, as well as complex reasoning logic such as modality corroboration, conflict resolution, and modality assistance, significantly increasing the difficulty of learning. This intrinsic complexity makes it costly and difficult to collect high-quality, fine-grained annotations. As a result, existing datasets typically contain only simple binary labels with limited modality coverage, lacking supervision for cross-modal relationships. As shown in Fig.~\ref{fig:fig1}, such weak supervision often causes models to overfit rigid patterns during SFT, while the RL phase suffers from insufficient feedback and exploration space, ultimately limiting the generalization and interpretability of SFT+RL, and even leading to worse performance than using SFT alone.

% 因此，我们首先系统性分析了当前多模态FAS数据集在SFT + RL训练范式下性能受限的核心原因，发现存在两大关键问题：(1) 监督微调阶段的多模态推理路径单一、任务类型单调、数据规模有限，弱化了对多模态信息的充分利用，严重压缩了强化学习阶段的探索空间，导致RL优化难以生效；(2) 即便通过传统的数据增强手段拓展输入多样性，模型仍可能出现仅依赖图像本身、忽略中间推理过程的“推理捷径”现象，即模型学习到一套对推理过程不敏感的hack策略，损害了可解释性与泛化能力。
% 为此，我们进一步探索了适用于低标注成本场景下的有效数据增强策略。在现有high-level增强方法基础上，我们提出**基于推理路径增强（Reasoning Path Augmentation）**的方法，以极低的代价显式扩展原始数据中的推理空间，从而显著提升RL阶段的探索能力与策略多样性。本文的主要贡献如下：
% (1)我们深入剖析了当前多模态FAS数据集在SFT + RL训练流程中的失效机制，为后续构建可解释、泛化能力强的训练范式提供了启示与实证依据；
% (2)为改善上述问题，我们提出一种新颖的推理路径增强策略，在有限标注下能够充分利用多模态信息推理且有效拓展RL阶段的推理空间，并且在SFT阶段引入打乱答案机制，让模型专注于推理路径的学习，避免推理捷径，提升了模型的泛化性与可解释性；
% (3)本工作首次系统整合多模态特征融合、域泛化能力提升与推理可解释性增强，构建了一个统一的、多维度能力协同的FAS框架。
Our analysis identifies two key issues in applying the SFT+RL paradigm to current multimodal FAS settings: 1) The SFT phase lacks multimodal reasoning diversity, involving simple tasks and limited-scale data, weakening the full utilization of multimodal information and severely narrowing the RL exploration space; and 2) Even with conventional data augmentation, models often exploit shortcuts by relying solely on image inputs, ignoring intermediate reasoning processes and leading to fragile decision strategies with poor generalization and interpretability. To address these issues, we propose a Reasoning Path Augmentation (PA) strategy, which explicitly expands the original reasoning space at minimal cost with positive–negative random path sampling method, enabling the full utilization of multimodal information for reasoning, as well as promoting greater exploration and policy diversity during the RL stage under limited label settings. Building upon high-level augmentation methods, PA introduces diverse multimodal reasoning chains associated with each input, thereby enhancing both generalization and interpretability. Our main contributions include:
\begin{itemize}
% \vspace{-0.8em}
\item We provide an in-depth analysis of the failure mechanisms of existing multimodal FAS datasets under the SFT+RL paradigm, offering both valuable insights and empirical evidence for designing explainable and generalizable training frameworks.
% \vspace{-0.2em}
\item We propose the PA-FAS framework with a novel Reasoning Path Augmentation strategy that fully utilization of multimodal information and effectively expands the reasoning space during the RL phase with limited supervision. Additionally, we introduce an answer shuffling mechanism during the SFT phase to focus the model on learning reasoning paths and avoid shortcuts, thereby enhancing the model's generalization and interpretability.
% \vspace{-0.2em}
\item To the best of our knowledge, this is the first work that systematically integrates multimodal feature fusion, domain generalization, and reasoning interpretability within a unified FAS framework, paving the way for robust and trustworthy multimodal FAS systems.
% \vspace{-1.3em}
\end{itemize}

\begin{figure}[t]
\vspace{-0.6em}
\centering
\includegraphics[width=0.38\textwidth]{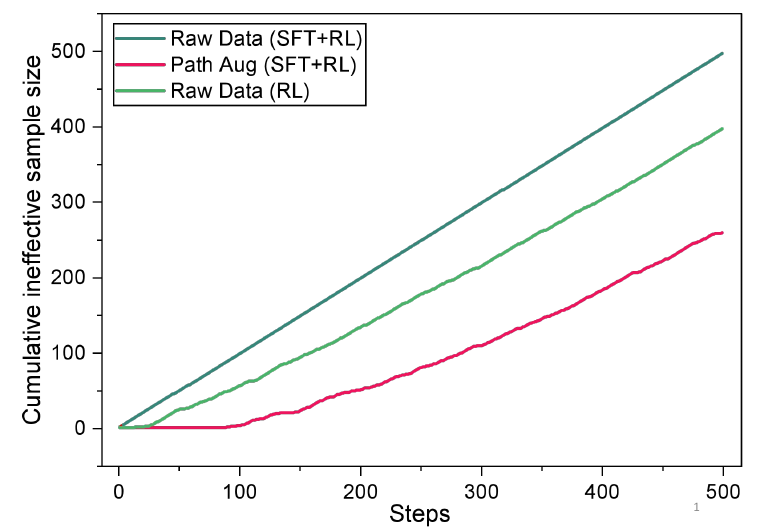}
\vspace{-1.1em}
% 在使用800种不同数据训练的模型中，强化学习（RL）阶段和监督微调加强化学习（SFT+RL）阶段的累积有效样本数量与训练步数的关系。在相同难度任务下，曲线上升的趋势可以间接衡量学习有效程度和探索空间大小
\caption{\small{Cumulative effective sample size versus training steps in the RL and SFT+RL stages for models trained with 800 data.}}
\label{fig:fail}
\vspace{-1.5em}
\end{figure}

\begin{figure*}[t]
\vspace{-2.0em}
\centering
\includegraphics[width=1.00\textwidth]{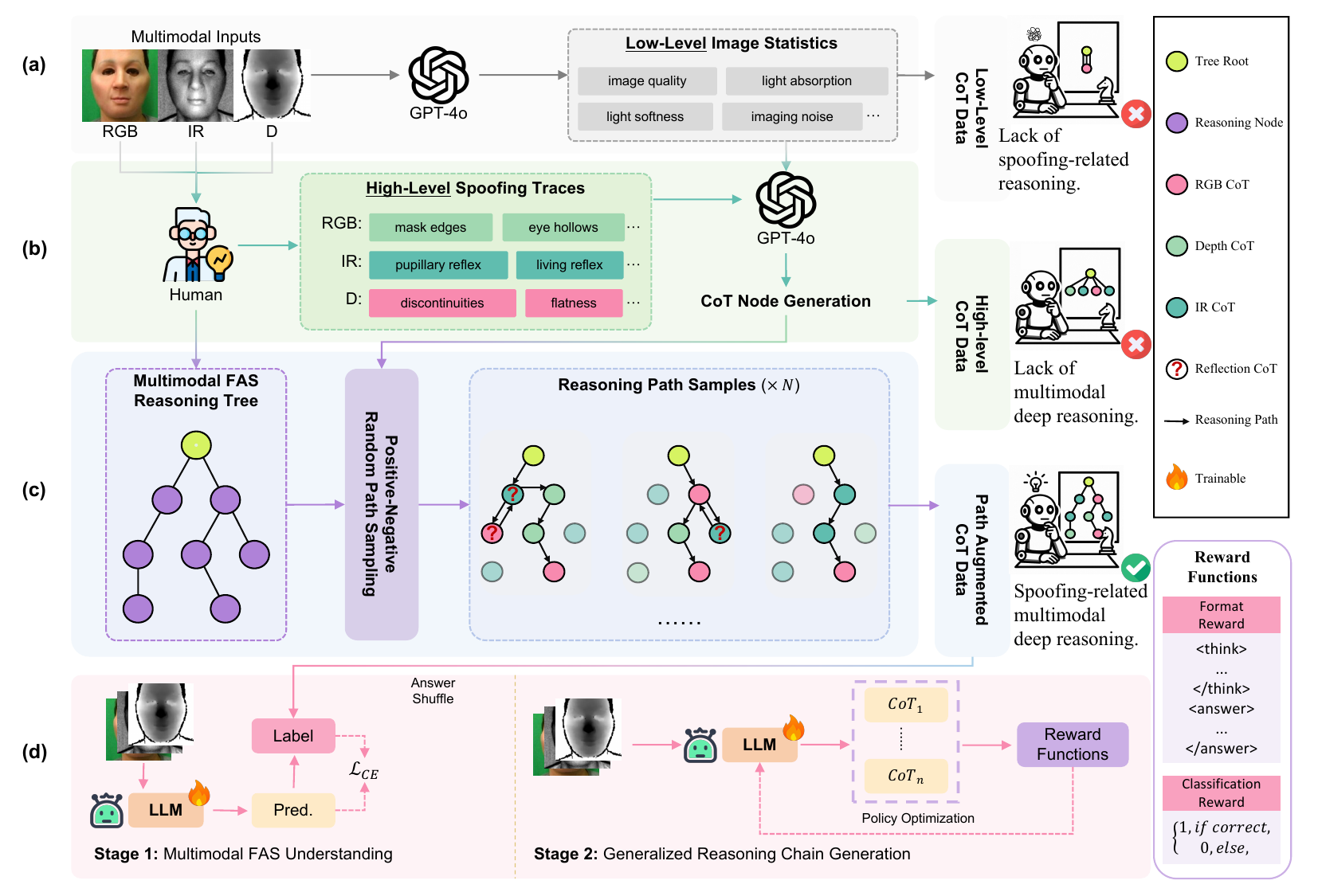}
% \vspace{-5.2em}
% PA-FAS框架示意图。原始数据（Raw Data）首先经过（a）low-level数据标注和（b）high-level数据标注得到对应的CoT。随后(c) 使用正负随机路径采样（Positive-Negative Random Path Sampling）方法，从人类构建的多模态推理树中采样指定数量的推理路径并组合成CoT。在采用（d）监督微调（SFT）+强化学习（RL）训练范式时，首先在SFT阶段对答案进行随机打乱，以避免策略模型形成思维捷径，从而学习多样化的推理路径和丰富的多模态领域特定知识。接着，在强化学习（RL）阶段，通过分类经历和格式化奖励，使策略模型逐步实现泛化。
\vspace{-2.6em}
\caption{\small{Schematic diagram of the PA-FAS framework. Raw data undergo (a) low-level and (b) high-level data annotation to obtain corresponding CoT. Subsequently, (c) Positive–Negative Random Path Sampling is employed to sample a specified number of reasoning paths from a human-constructed multimodal reasoning tree and integrate them into CoT. During the (d) SFT+RL training paradigm, answers are randomly shuffled in the SFT stage to prevent the policy model from forming shortcuts, thereby learning diverse reasoning paths and rich multimodal domain-specific knowledge. In the RL stage, the policy model achieves generalization through classification and format rewards.}}
\label{fig:framework}
\vspace{-1.1em}
\end{figure*}

\vspace{-0.6em}
\section{Related Works}
%\subsection{Multimodal Generalized Face Anti-Spoofing}

%\vspace{0.2em}
\noindent\textbf{Multimodal Face Anti-Spoofing.} \quad   With the advancement of deep learning, numerous FAS methods based on CNNs and ViTs have been proposed \cite{jiang2023adversarial,yue2023cyclically,liu2023towards,cai2023rehearsal,sadapter2024,camoeit2024}. Domain generalization (DG) aims to train models on multiple source domains to generalize well to unseen targets. Recent approaches have achieved promising results by learning cross-domain shared spaces \cite{safas2023}, domain-invariant features \cite{liao2023domain,iadg2023,liu2024cfpl}, and incorporating textual prompts \cite{flipfas2023,liu2024cfpl,vlfas2024}. However, these single-modal methods often overlook modality bias, limiting their effectiveness in multimodal scenarios. In contrast, multimodal FAS methods \cite{yu2023flexible,fmvit2023,han2023hyperbolic,m3fas2024,ama2024,vpfas2024} leverage complementary cues from sensors such as RGB, DEPTH, and Infrared to better detect spoofing. To enhance cross-modal learning, some works introduce attention-based fusion and adaptive loss functions \cite{li2023learning}, while others use cross-modal translation to reduce semantic gaps \cite{li2021asymmetric}. Nonetheless, most of these approaches ignore domain shift, hindering generalization. To bridge this gap, recent studies have begun to jointly address multimodal learning and domain generalization, exploring strategies \cite{MMDG2024,dadm2025,bigmoe2025,DBLP:journals/pami/LinLYCWYWLCK25} such as uncertainty-aware adapter and dual alignment. Despite these advances, limited interpretability remains a key obstacle to practical deployment.

%\subsection{MLLMs for Face Anti-Spoofing}
%\vspace{0.2em}
\noindent\textbf{MLLMs for Face Anti-Spoofing.} \quad   The rapid development of MLLM highlights significant progress in processing intertwined visual and textual data. Models such as GPT-4V \cite{gpt4v2023}, Qwen2.5VL \cite{qwen2_5_2025}, and Gemini \cite{gemini2023} have achieved remarkable breakthroughs in various vision-related tasks. Concurrently, reinforcement learning approaches like GRPO \cite{grpo2024} have notably improved computational efficiency and enhanced long Chain-of-Thought (CoT) task performance. Recent studies \cite{samr12025,deepperception2025,r1omni2025,vrft2025,medvlmr12025} demonstrate that incorporating reinforcement learning into MLLM training significantly boosts performance and generalization across a wide range of vision tasks. Given the importance of knowledge integration and reasoning format for domain-specific downstream applications, the combined use of SFT Memorizes, RL Generalizes strategies \cite{chu2025sft} has gained substantial adoption. In the FAS domain, several works have explored the potential of MLLMs for domain-specific tasks \cite{zhang2025interpretable,shield2025,facecot2025}, showcasing their strong interpretability and multitask capabilities. However, research specifically addressing multimodal domain generalization in FAS using MLLMs remains scarce. The performance boundaries and associated challenges of such applications still warrant further investigation and in-depth study.

\vspace{-0.7em}
\section{Method}
\subsection{Analysis of Failure in Multimodal FAS Datasets}
\vspace{-0.1em}
% 已有研究表明，相较于单独使用监督微调（Supervised Fine-Tuning, SFT）或强化学习（Reinforcement Learning, RL），将SFT与RL相结合的训练范式能够在多个任务中显著提升性能 \cite{chu2025sft}。其关键在于，SFT阶段帮助模型掌握特定领域的知识、推理路径及思维链格式，为RL阶段的策略优化提供基础。然而，在多模态FAS任务中直接应用该范式时，由于现有数据集通常仅包含图像及对应的二分类标签，缺乏对关键视觉线索的语言标注，且任务类型高度单一，模型容易在SFT阶段形成对输出结果的过度置信。这种过度置信导致模型在RL阶段几乎始终获得极端奖励（如全部为1或全部为0），缺乏中间反馈信号。如图~\ref{fig:fail}所示，使用原始数据进行SFT训练的模型在RL阶段快速累积reward=1的无效样本数，曲线近似线性上升，表明模型缺乏对策略空间的有效探索。而直接使用RL训练的模型虽未经过SFT，其累积曲线则增长更为平缓，反映出一定的探索性。这一现象突显了现有数据设置下SFT+RL范式的局限性
Previous studies demonstrate that combining SFT with RL can significantly improve model performance across various tasks \cite{chu2025sft}, as the SFT stage allows the model to internalize domain-specific knowledge, reasoning patterns, and chain-of-thought structures, laying the groundwork for effective policy optimization during RL. However, when this paradigm is applied to multimodal FAS tasks, where datasets with only binary labels, lacking linguistic annotations of key visual cues, and exhibiting high task uniformity, the model often develops overconfident predictions during the SFT phase. Such overconfidence leads to extreme reward feedback during RL, where most samples receive either full (1) or zero rewards, and informative intermediate signals are largely absent. As shown in Fig.~\ref{fig:fail}, the model fine-tuned on raw data accumulates ineffective samples at a nearly linear rate throughout RL, indicating a lack of exploration and highly polarized learning signals. In contrast, the model trained directly with RL, although not fine-tuned, demonstrates a more moderate cumulative trajectory, suggesting greater exploratory behavior. These findings highlight the limitations of the SFT+RL paradigm under current dataset settings and motivate the need for improved training strategies in the fine-tuning stage.

\begin{figure}[t]
\vspace{-1.5em}
\centering
\includegraphics[width=0.49\textwidth]{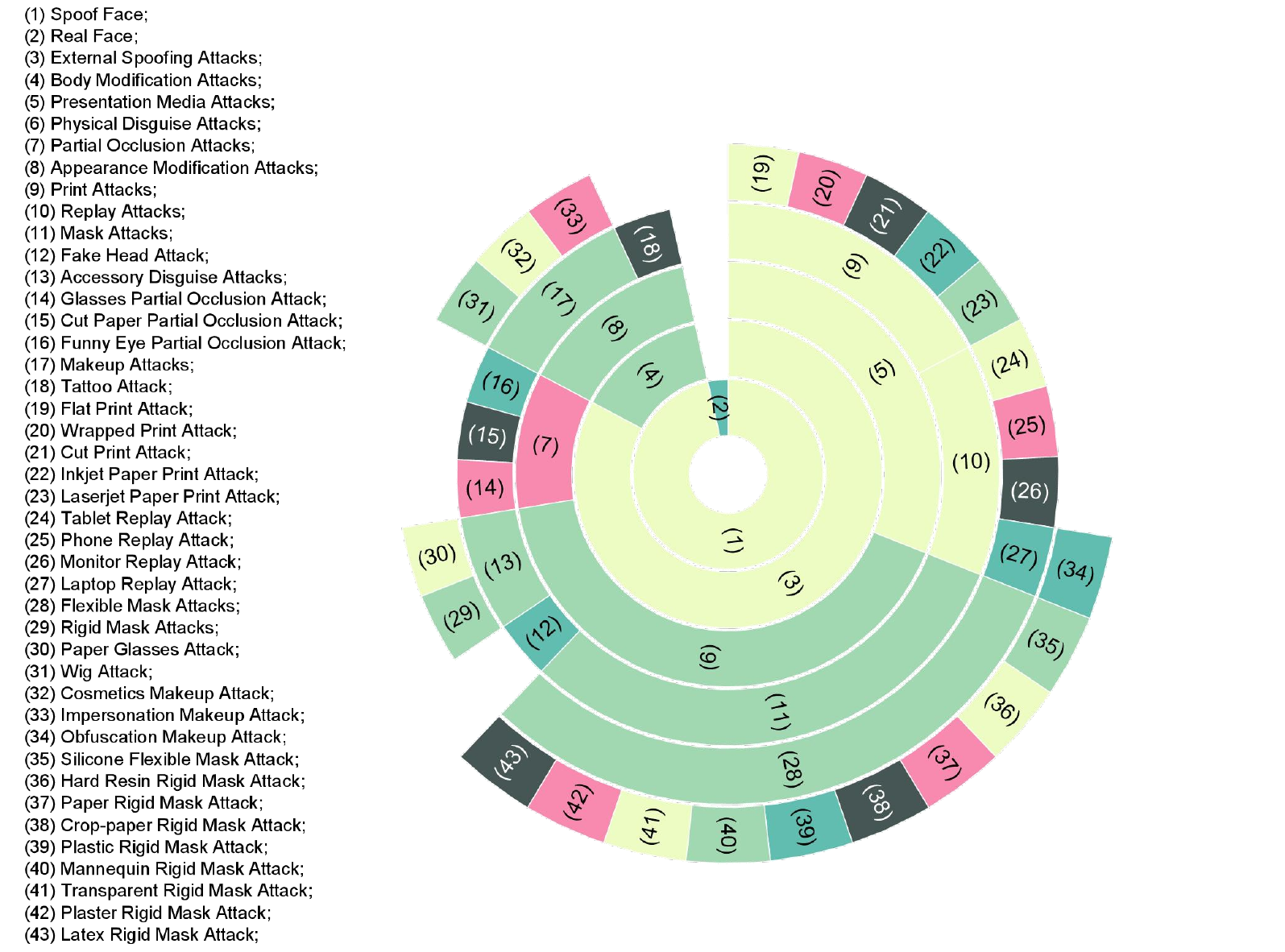}
\vspace{-1.7em}
% 人脸防欺骗（FAS）细粒度层次分类的Sunburst图，其中每个分类都被直接用作推理树的节点。
\caption{\small{Sunburst diagram of the fine-grained hierarchical taxonomy for FAS. Every category is directly mapped to a node in the reasoning tree.}}
\label{fig:reason tree}
\vspace{-1.4em}
\end{figure}

\begin{algorithm}[t]\small
    \caption{\small Positive–Negative Random Path Sampling}
    \label{alg:pnrps}
    \KwIn{classification tree $T$ with root node $r$; target leaf label $\ell$; length parameter $\alpha$; sampling number $N$}
    \KwOut{sampled paths $P$}
    
    \textcolor{teal}{\small \# Compute maximum path length}\\
    $D \leftarrow$ maximum depth of $r$; \quad $L_{\text{max}} \leftarrow \alpha(D - 1)$\;
    
    \textcolor{teal}{\small \# Initialize samples set and exploration stack}\\
    $P \leftarrow \emptyset$; \quad $S \leftarrow \{(r,\,[\,])\}$\;
    
    \textcolor{teal}{\small \# Begin depth-first positive‐negative sampling}\\
    \While{$S \neq \emptyset$}{
        \textcolor{teal}{\small \# Stop if enough samples collected}\\
        \If{$|P| \ge N$}{
            \textbf{break}\;
        }

        \textcolor{teal}{\small \# Pop next node and its path}\\
        Pop $(node,\,path)$ from $S$\;
        \If{$|path| \ge L_{\text{max}}$}{
            \textbf{continue}\;
        }
        \If{$node.name = \ell$}{
            append $path$ to $P$\;
            \textbf{continue}\;
        }
        
        \textcolor{teal}{\small \# Positive exploration (forward step)}\\
        \For{$child$ \textbf{in} $node.children$ \textbf{if} $child.^{+} = \text{false}$}{
            $child.^{+} = \text{true}$\;
            push $(child,\,path \cup\{(+,child.name)\})$ onto $S$\;
        }
        
        \textcolor{teal}{\small \# Negative reflection (backward step)}\\
        \If{$node \neq r$ \textbf{and} $node.^{-} = \text{false}$}{
            $node.^{-} = \text{true}$\;
            push $(node.parent,\,path \cup\{(-,node.name)\})$ onto $S$\;
        }
    }
    
    \textcolor{teal}{\small \# Return sampled paths}\\
    \Return $P$\;
\end{algorithm}

\begin{table*}[t]
 \centering
 \vspace{-0.8em}
\caption{\small{Cross-dataset testing results under the fixed-modal scenarios (Protocol 1) among CASIA-CeFA (C), PADISI (P), CASIA-SURF (S), and WMCA (W). Best and second-best results are marked in \textbf{bold} and \underline{underline}, respectively.} }
 \vspace{-0.8em}
    \resizebox{1.00\textwidth}{!}{
    \begin{tabular}{l|cc|cc|cc|cc|cc} 
        \toprule[1.3pt]
        \multirow{2}{*}{\textbf{Method}}& \multicolumn{2}{c|}{\textbf{CPS$\rightarrow$W}}&\multicolumn{2}{c|}{\textbf{CPW$\rightarrow$S}}&\multicolumn{2}{c|}{\textbf{CSW$\rightarrow$P}}&\multicolumn{2}{c|}{\textbf{PSW$\rightarrow$C}}&\multicolumn{2}{c}{\textbf{Average}}\\
        \cmidrule(r){2-3} \cmidrule(r){4-5} \cmidrule(r){6-7} \cmidrule(r){8-9} \cmidrule(r){10-11}
        & HTER(\%)$\downarrow$ & AUC(\%)$\uparrow$ & HTER(\%)$\downarrow$ & AUC(\%)$\uparrow$& HTER(\%)$\downarrow$ & AUC(\%)$\uparrow$ & HTER(\%)$\downarrow$ & AUC(\%)$\uparrow$& HTER(\%)$\downarrow$ & AUC(\%)$\uparrow$ \\
        \midrule
        \multicolumn{11}{c}{\textbf{Uni-modal DG (Concat + 1*1 Conv)}}\\
        \midrule
        SSDG \textcolor{gray}{[CVPR'20]}&26.09&82.03&28.50&75.91&41.82&60.56&40.48&62.31&37.32&68.25\\
        SSAN \textcolor{gray}{[CVPR'22]}&17.73&91.69&27.94&79.04&34.49&68.85&36.43&69.29&35.34&70.98\\
        SA-FAS \textcolor{gray}{[CVPR'23]}&21.37&87.65&23.22&84.49&35.10&70.86&35.38&69.71&28.77&78.18\\
        IADG \textcolor{gray}{[CVPR'23]}&27.02&86.50&23.04&83.11&32.06&73.83&39.24&63.68&39.83&62.95\\
        FLIP \textcolor{gray}{[IJCAI'22]}&13.19&93.79&\underline{11.73}&\underline{94.93}&17.39&90.63&22.14&83.95&16.11&\underline{90.83}\\
        \midrule
        \multicolumn{11}{c}{\textbf{Multi-modal FAS}}\\
        \midrule
        ViT \textcolor{gray}{[ICLR'20]}&20.88&84.77&44.05&57.94&33.58&71.80&42.15&56.45&36.60&68.12\\
        AMA \textcolor{gray}{[IJCV'24]}&17.56&88.74&27.50&80.00&21.18&85.51&47.48&55.56&27.47&79.85\\
        VP-FAS \textcolor{gray}{[TDSC'24]}&16.26&91.22&24.42&81.07&21.76&85.46&39.35&66.55&29.82&76.62\\
        ViTAF \textcolor{gray}{[ECCV'22]}&20.58&85.82&29.16&77.80&30.75&73.03&39.75&63.44&33.89&71.54\\
        MM-CDCN \textcolor{gray}{[CVPR'20]}&38.92&65.39&42.93&59.79&41.38&61.51&48.14&53.71&46.81&53.43\\
        CMFL \textcolor{gray}{[CVPR'21]}&18.22&88.82&31.20&75.66&26.68&80.85&36.93&66.82&31.01&75.07\\
        CLIP \textcolor{gray}{[ICML'21]}&14.55&90.47&18.17&90.02&24.13&83.15&38.33&65.71&24.63&83.00\\
        \midrule
        \multicolumn{11}{c}{\textbf{Multi-modal DG FAS}}\\
        \midrule
        MMDG \textcolor{gray}{[CVPR'24]}&12.79&93.83&15.32&92.86&18.95&88.64&29.93&76.52&22.93&84.19\\
        % MMDG++ \cite{mmdgplus2025} &2.08&99.82&8.72&96.77&10.24&94.97&18.87&89.28&9.98&95.21\\
        DADM \textcolor{gray}{[ICCV'25]}&11.71&94.89&\textbf{6.92}&\textbf{97.66}&19.03&88.22&\textbf{16.87}&\textbf{91.08}&\textbf{13.63}&\textbf{92.96}\\
        % MMDA \cite{mmda2025}&1.22&99.99&4.21&98.62&4.34&98.58&6.25&98.18&4.00&98.94\\
        \midrule
        \multicolumn{11}{c}{\textbf{Interpretable Multi-modal DG FAS}}\\
        \midrule
        Qwen2.5-VL-3B \textcolor{gray}{[ARXIV'25]}&30.86&75.01&49.56&44.35&19.72&88.1&33.72&70.01&33.46&69.36\\
        Qwen2.5-VL-3B-SFT&\underline{5.12}&\underline{97.65}&44.84&52.71&\underline{15.15}&\underline{90.73}&27.91&76.20&23.25&79.32\\
        Qwen2.5-VL-3B-SFT+GRPO&7.75&97.36&57.06&44.79&45.26&56.29&27.41&76.29&34.37&68.68\\
        \textbf{PA-FAS (Ours)}&\textbf{2.39}&\textbf{99.73}&27.75&78.07&\textbf{9.48}&\textbf{94.5}&\underline{21.23}&\underline{84.25}&\underline{15.21}&89.13\\
        \bottomrule[1.3pt]
    \end{tabular}
    }
     
    \label{table:P1}
    \vspace{-1.3em}
\end{table*}

% 为缓解上述问题，我们首先尝试了一种冷启动数据增强策略。鉴于为多模态图像手动标注关键特征线索的成本较高，我们采用了较为自然的替代方案：针对少量已标注样本中的每个样本，构建多个不同版本的Chain-of-Thought（CoT）推理链，以引入推理路径的多样性并提升模型的泛化能力。然而该基于High-level级的增强策略未能有效缓解极端奖励分布的问题。进一步地，我们设计了思维内容替换实验：在SFT阶段将<think>标签内的推理文本随机替换为其他样本的推理过程。实验结果（图 \ref{fig:fig1}(a)）显示，无论是否替换，模型在SFT和RL阶段的表现几乎一致，表明模型并未真正利用CoT内容进行推理，而是直接依赖图像输入给出答案，形成所谓的“推理捷径”。因此，我们认为要实现SFT+RL在多模态FAS任务中的有效训练，需满足两个关键条件：(1) 在SFT阶段提供结构丰富且具有多样性的推理路径，以构建有效的策略搜索空间；(2) 防止模型在SFT阶段形成对图像与答案之间的直接映射，避免忽视思维链本身的学习价值。
% 为满足第一个条件，我们提出了**推理路径增强（Reasoning Path Augmentation, PA）**策略。该方法不同于High-level级数据增强，而是着眼于推理链本身的结构多样性扩展。通过为每个样本构建多个具备语义一致性但推理逻辑差异的CoT路径，如图\ref{fig:fail}所示，PA显著扩展了RL阶段的可用探索空间，在有限标注条件下实现了推理路径的丰富化和泛化能力的提升。为解决第二个问题，我们在 SFT 阶段引入“答案随机打乱”这一极简机制，迫使模型同时学习多样化的推理路径和所有潜在答案；此举不仅阻断推理捷径，还为 RL 阶段保留了更广阔的答案探索空间。
To mitigate this issue, we first adopt a cold-start data augmentation strategy. Considering the high cost of manually annotating key visual clues in multimodal images, we pursue a more practical approach: as shown in Fig.~\ref{fig:framework}~(a) and (b), for each of the few annotated samples, we generate multiple distinct versions of low-level and high-level CoT reasoning chains, aiming to introduce diversity in reasoning paths and enhance the model's generalization capability. However, as illustrated in Fig.~\ref{fig:fig1}~High-level Aug (800), this augmentation fails to alleviate the problem of extreme reward distribution. To further investigate, we conduct a reasoning content replacement experiment, wherein the reasoning text enclosed by $<think>$ tags during SFT is randomly substituted with CoT sequences from other samples. As shown in Fig.~\ref{fig:fig1}~`High-level Aug + Path Shuffle (800)', the performance of the model remains nearly unchanged regardless of whether the reasoning content is replaced, indicating that reasoning heavily relies on the visual predictions but neglects the CoT, a phenomenon we refer to as `reasoning shortcut'. Therefore, to enable effective training under the SFT+RL paradigm for multimodal FAS tasks, we identify two key requirements: 1) the SFT stage must provide structurally diverse and semantically rich reasoning paths to construct a meaningful exploration space for RL; and 2) shortcut learning based on direct image-to-answer mapping should be avoided to ensure the model learns to reason explicitly through the CoT process. To satisfy the first condition, we propose the PA-FAS framework with a Reasoning Path Augmentation (PA) strategy. Unlike high-level data augmentation, PA focuses on diversifying the structure of reasoning chains. By constructing multiple CoT paths for each sample that are semantically consistent but logically varied, as shown in Fig.~\ref{fig:fail}, PA significantly expands the model’s exploration space during RL. This enables effective reasoning path generalization with only limited annotated data. To address the second issue, we introduce an answer-shuffling mechanism during SFT, compelling the model to master diverse reasoning paths and every possible answer, which effectively blocks reasoning shortcuts while reserving ample exploration space for the RL phase.

\subsection{Reasoning Path Augmentation}

% 我们提出的推理路径增强策略（Reasoning Path Augmentation, PA）旨在通过构造结构化、多样化的推理路径集合}，拓展模型在强化学习阶段的输出空间，从而显著提升其探索能力与训练效率。在数据资源有限的条件下，high-level 级数据增强虽然可以引入一定程度的语义多样性，但其对有效推理路径的扩展极为稀疏。同时，As illustrated in Figure 1, the effectiveness of High-level augmentation shows a significant decline compared to Low-level augmentation，这是由于在多模态 FAS任务中，利用现有多模态大语言模型进行数据增强往往代价高昂，难以有效过滤语义错误或逻辑冲突的伪增强文本，使得增强数据往往由于大量错误和噪声显著降低其数据质量，因此难以在大规模场景中推广应用。
As shown in Fig.~\ref{fig:framework}, we propose a novel PA strategy that constructs a structured and diverse set of reasoning paths to expand the output space during reinforcement learning, thereby significantly enhancing both exploration efficiency and training effectiveness. As illustrated in Fig.~\ref{fig:fig1}, the effectiveness of High-level augmentation shows a significant decline compared to Low-level augmentation, because under limited data availability, data augmentation by existing MLLM can introduce some degree of semantic diversity but remains extremely sparse in expanding valid reasoning trajectories. Moreover, in multimodal FAS tasks, token-level operations are often prohibitively costly and fail to effectively filter out semantically erroneous or logically inconsistent pseudo-augmented texts, causing the augmented data to suffer from substantial noise and errors that markedly degrade data quality and thus hinder large-scale deployment.

% 为了解决上述问题，我们提出一种基于推理路径的数据增强方法。该方法以如图\ref{fig:reason tree}所示的多模态FAS任务中细粒度的标签层级关系为基础，构建形式化的推理分类树 $\mathcal{T} = (\mathcal{V}, \mathcal{E})$，其中 $\mathcal{V}$ 为推理节点集合，$\mathcal{E}$ 为有向边集合。每个节点 $v \in \mathcal{V}$ 表示一个语义类别或逻辑判断单元，每条路径 $\mathcal{P} = (v_1, v_2, \dots, v_n)$ 对应一条从根节点 $v_1 = r$ 到目标叶节点 $v_n = \ell$ 的完整推理链。
To address these challenges, we propose a reasoning-path-based data augmentation method. Leveraging the fine-grained label hierarchy illustrated in Fig.~\ref{fig:reason tree} for multimodal FAS tasks, we build a formal reasoning tree $\mathcal{T} = (\mathcal{V}, \mathcal{E})$, where $\mathcal{V}$ is the set of reasoning nodes and $\mathcal{E}$ the set of directed edges. Each node $v \in \mathcal{V}$ represents a semantic class or logical decision unit, and every path $\mathcal{P} = (v_1, v_2, \dots, v_n)$ encodes a complete reasoning chain from the root $v_1 = r$ to the target leaf node $v_n$.

% 在此基础上，我们提出正反向随机路径采样（Positive--Negative Random Path Sampling）策略（见算法~\ref{alg:pnrps}），用于系统地采样出覆盖广泛、逻辑多样的推理路径集 $\mathcal{P}_i = \{\mathcal{P}_i^{(1)}, \dots, \mathcal{P}_i^{(k)}\}$，以增强每个原始样本 $x_i$ 的推理表示能力。该方法的核心机制包括以下几点：(1) 单节点操作约束：每个节点 $v \in \mathcal{V}$ 最多执行一次\textit{正向探索}操作（forward step: $(+, v)$）和一次\textit{反向反思}操作（backward step: $(-, v)$），以避免冗余游走；(2) 路径长度限制：设分类树最大深度为 $D$，则路径最大长度上限为 $L_{\text{max}} = \alpha(D - 1)$，以控制生成路径的复杂度,此处$\alpha>1$是控制参数；(3) 语义一致性约束：路径中每个节点对应一个预定义的 Chain-of-Thought（CoT）子句模板，最终推理文本通过路径序列拼接这些模板构成；(4) 结构采样策略：采用深度优先搜索从根节点出发，在保证语义有效性的前提下，随机采样多条利用任意模态的合法路径，并映射生成新的增强文本。我们将推理路径增强过程形式化为以下映射：
% \begin{equation}
%     \text{PA}: \quad \{(x_i, \ell_i)\}_{i=1}^N \rightarrow \bigcup_{i=1}^N \left\{ \left(x_i,\, \text{CoT}(\mathcal{P}_i^{(j)})\right) \;\middle|\; \mathcal{P}_i^{(j)} \in \mathcal{T},\; j = 1, \dots, k \right\},
% \end{equation}
% 其中 $\text{CoT}(\mathcal{P})$ 表示由路径 $\mathcal{P}$ 对应的推理单元组合生成的逻辑链条描述。
Built upon this structure, we propose a Positive--Negative Random Path Sampling (PNRPS) strategy (see Algorithm~\ref{alg:pnrps}) to systematically sample a diverse and logically rich set of reasoning paths $\mathcal{P}_i = \{\mathcal{P}_i^{(1)}, \dots, \mathcal{P}_i^{(N)}\}$ for each original data $x_i$, thereby enhancing its reasoning ability. Assume a dataset consisting of $M$ instances $\{(x_i, \ell_i)\}_{i=1}^M$, where each $x_i$ is a data sample and $\ell_i$ is its corresponding ground-truth label. The core mechanisms are as follows: (1) Single-node operation constraint: Each node $v \in \mathcal{V}$ is allowed at most one forward exploration step $(+, v)$ and one backward reflection step $(-, v)$ to avoid redundant walks; (2) Path length constraint: Given a maximum taxonomy depth $D$, we set an upper bound $L_{\text{max}} = \alpha(D - 1)$ on the reasoning path length to control complexity, where $\alpha > 1$ is a tunable scaling factor that determines the maximum allowable reasoning steps relative to the taxonomy depth; (3) Semantic consistency enforcement: Each node is associated with a predefined Chain-of-Thought (CoT) clause template, and the final reasoning text is constructed by sequentially composing these templates along the path; (4) Structural sampling strategy: We perform rule-guided depth-first traversal from the root, randomly sampling $N$ valid paths that utilize information from any of the RGB, IR, and DEPTH modalities and mapping them into logically coherent CoT descriptions. The PA process can be formalized as a mapping:
{\small
\vspace{-0.6em}
\begin{equation}
\begin{aligned}
    \{(x_i, \ell_i)\}_{i=1}^M \rightarrow
    \bigcup_{i=1}^M \Big\{ \left(x_i,\, \text{CoT}(\mathcal{P}_i^{(j)})\right)\, \Big|\, j = 1, \dots, N \Big\}
\end{aligned}
\end{equation}
}
where $\text{CoT}(\mathcal{P})$ represents the reasoning description composed by chaining CoT sub-clauses corresponding to path $\mathcal{P}$.

% 在实际实验中，我们仅基于约 800 条原始标注样本，对每个样本采样 50 条推理路径，即可生成约 $4 \times 10^4$ 条语义一致、结构清晰的增强样本。该方法极大提升了数据的\textbf{多样性覆盖率}与\textbf{可控性结构性}，为后续强化学习提供了更稳健且具有泛化能力的训练支撑。
In our implementation, we start with only 800 labeled instances and generate $N=50$ reasoning paths per data, yielding a total of approximately $4 \times 10^4$ structurally diverse and semantically coherent augmented samples. This strategy significantly improves the coverage diversity and structural controllability of the training data, providing a robust foundation for reinforcement learning with improved generalization and reasoning capability.

\begin{table*}[t]
    \centering
    \vspace{-0.8em}
    \caption{\small{Cross-dataset testing results under the missing modalities scenarios (Protocol 2) among CASIA-CeFA (C), PADISI (P), CASIA-SURF (S), and WMCA (W). Best and second-best results are marked in \textbf{bold} and \underline{underline}, respectively.}} 
    \vspace{-0.9em}
    \tabcolsep=4mm
    \resizebox{1.00\textwidth}{!}{
    \begin{tabular}{l|cc|cc|cc|cc} 
        \toprule[1.3pt]
        \multirow{2}{*}{\textbf{Method}}& \multicolumn{2}{c|}{\textbf{Missing D}}&\multicolumn{2}{c|}{\textbf{Missing I}}&\multicolumn{2}{c|}{\textbf{Missing D \& I}}&\multicolumn{2}{c}{\textbf{Average}}\\
        \cmidrule(r){2-3} \cmidrule(r){4-5} \cmidrule(r){6-7} \cmidrule(r){8-9}
        & HTER(\%)$\downarrow$ & AUC(\%)$\uparrow$ & HTER(\%)$\downarrow$ & AUC(\%)$\uparrow$& HTER(\%)$\downarrow$ & AUC(\%)$\uparrow$ & HTER(\%)$\downarrow$ & AUC(\%)$\uparrow$\\
        \midrule
        \multicolumn{9}{c}{\textbf{Uni-modal DG (Concat + 1*1 Conv)}}\\
        \midrule
        SSDG \textcolor{gray}{[CVPR'20]}&38.92&65.45&37.64&66.57&39.18&65.22&38.58&65.75\\
        SSAN \textcolor{gray}{[CVPR'22]}&36.77&69.21&41.20&61.92&33.52&73.38&37.16&68.17\\
        SA-FAS \textcolor{gray}{[CVPR'23]}&36.30&69.07&39.80&62.69&33.08&74.29&36.40&68.68\\
        IADG \textcolor{gray}{[CVPR'23]}&40.72&58.72&42.17&61.83&37.50&66.90&40.13&62.49\\
        FLIP \textcolor{gray}{[IJCAI'22]}&23.66&83.90&24.06&84.04&27.07&79.79&27.93&79.44\\
        \midrule
        \multicolumn{9}{c}{\textbf{Multi-modal FAS}}\\
        \midrule
        ViT \textcolor{gray}{[ICLR'20]}&40.04&64.69&36.77&68.19&36.20&69.02&37.67&67.30\\
        AMA \textcolor{gray}{[IJCV'24]}&29.25&77.70&32.30&74.06&31.48&75.82&31.01&75.86\\
        VP-FAS \textcolor{gray}{[TDSC'24]}&29.13&78.27&29.63&77.51&30.47&76.31&29.74&77.36\\
        ViTAF \textcolor{gray}{[ECCV'22]}&34.99&73.22&35.88&69.40&35.89&69.61&35.59&70.64\\
        MM-CDCN \textcolor{gray}{[CVPR'20]}&44.90&55.35&43.60&58.38&44.54&55.08&44.35&56.27\\
        CMFL \textcolor{gray}{[CVPR'21]}&31.37&74.62&30.55&75.42&31.89&74.29&31.27&74.78\\
        CLIP \textcolor{gray}{[ICML'21]}&28.07&77.00&29.10&77.04&32.58&73.36&33.83&71.11\\
        \midrule
        \multicolumn{9}{c}{\textbf{Multi-modal DG FAS}}\\
        \midrule
        MMDG \textcolor{gray}{[CVPR'24]}&24.89&82.39&23.39&83.82&25.26&81.86&24.51&82.69\\
        % MMDG++ \cite{mmdgplus2025} &15.11&92.01&15.56&91.05&17.64&89.51&16.10&90.85\\
        DADM \textcolor{gray}{[ICCV'25]}&\underline{21.56}&\underline{85.17}&\underline{20.82}&\underline{85.28}&\underline{22.61}&\underline{84.04}&\underline{21.66}&\underline{84.83}\\
        % MMDA \cite{mmda2025}&11.10&93.97&5.98&98.30&13.36&93.74&10.14&95.33\\
        \midrule
        \multicolumn{9}{c}{\textbf{Interpretable Multi-modal DG FAS}}\\
        \midrule
        Qwen2.5-VL-3B \textcolor{gray}{[ARXIV'25]}&33.46&69.36&33.46&69.36&33.46&69.36&33.46&69.36\\
        Qwen2.5-VL-3B-SFT&23.25&79.32&23.25&79.32&23.25&79.32&23.25&79.32\\
        Qwen2.5-VL-3B-SFT+GRPO&34.37&68.68&34.37&68.68&34.37&68.68&34.37&68.68\\
        \textbf{PA-FAS (Ours)}&\textbf{15.68}&\textbf{89.07}&\textbf{17.32}&\textbf{88.23}&\textbf{14.67}&\textbf{89.73}&\textbf{15.85}&\textbf{89.01}\\
        \bottomrule[1.3pt]
    \end{tabular}
    }
    \vspace{-1.0em}
    
    \label{table:P2}
\end{table*}

\subsection{Answer Shuffling for SFT+RL Paradigm}

% 尽管我们引入了推理路径增强机制以提升训练样本的多样性与规模，但由于单一任务结合丰富的多模态推理路径会导致SFT阶段对推理过程产生一定的困惑，这会诱发答案部分的生成直接参考图像输入，忽略了思考过程，形成推理捷径。这同样也会导致经过SFT阶段训练后的模型过于自信，极大的压缩了模型在RL阶段的探索空间。因此为了切断推理捷径，我们在图\ref{fig:framework}~(d) SFT阶段引入了answer shuffling机制，该机制会将思维链中的最终答案与其他数据样本中的答案进行随机替换,如此一来促进模型专心学习推理过程中的丰富的推理路径，同时也为所有可能的答案保留探索空间。

% 经过answer shuffling机制的SFT阶段后，在图\ref{fig:framework}~(c) RL 阶段，我们采用了 Group Relative Policy Optimization（GRPO）方法，其核心思想是利用组内相对优势替代传统值函数，从而更稳定地指导策略优化。具体而言，对于一个给定的问题-答案对 \((q, a)\)，旧策略 \(\pi_{\theta_\text{old}}\) 会采样一组 \(G\) 个响应 \(\{o_i\}_{i=1}^G\)，并基于其对应奖励 \(\{r_i\}_{i=1}^G\) 计算相对优势如下：
% \begin{equation}
%     \hat{A}_{i,t} = \frac{r_i - \mathrm{mean}(\{R_i\}_{i=1}^G)}{\mathrm{std}(\{R_i\}_{i=1}^G)}.
%     \label{eq:relative_advantage}
% \end{equation}
% GRPO 的优化目标函数如下所示，引入了 clipped 概率比和 KL 正则项：
% \begin{equation}
% \begin{aligned}
% \mathcal{J}_{\text{GRPO}}(\theta) = \mathbb{E}_{(q,a)\sim\mathcal{D}, \{o_i^G\}\sim\pi_{\theta_{\text{old}}}} \Bigg[ \frac{1}{G} \sum_{i=1}^G \frac{1}{|o_i|} \sum_{t=1}^{|o_i|} \Big( &\min \big( r_{i,t}(\theta) \hat{A}_{i,t},\; \mathrm{clip}(r_{i,t}(\theta), 1 - \varepsilon, 1 + \varepsilon)\hat{A}_{i,t} \big) \\
% & - \beta D_\text{KL}(\pi_\theta \| \pi_{\text{ref}}) \Big) \Bigg],
% \end{aligned}
% \label{eq:grpo_objective}
% \end{equation}
% 其中，\(r_{i,t}(\theta)\) 表示当前策略与旧策略在时间步 \(t\) 的比值，计算方式为：
% \begin{equation}
%     r_{i,t}(\theta) = \frac{\pi_\theta(o_{i,t} \mid q, o_{i,<t})}{\pi_{\theta_{\text{old}}}(o_{i,t} \mid q, o_{i,<t})}.
%     \label{eq:ratio}
% \end{equation}
% 在原始 GRPO 形式中，KL 散度项主要用于限制策略模型的更新幅度，防止其过度偏离旧策略。然而，这种设计在数据稀缺条件下可能进一步压制策略模型的探索能力,因此我们弃用了KL散度项。
Although reasoning-path augmentation enlarges the training set, the coupling of a single task with overly rich multimodal reasoning paths can confuse the model during SFT: the final answer is often produced by directly looking at the image, bypassing the reasoning chain and creating a shortcut. This shortcut makes the model overconfident and drastically shrinks the exploration space in the RL phase. To sever such shortcuts, we introduce answer shuffling in Fig.~\ref{fig:framework}(d). During SFT, the final answer in each chain-of-thought is randomly swapped with the answer from another sample, forcing the model to focus on learning the diverse reasoning paths instead of memorizing the answer and preserving room to explore every possible answer.

After the answer-shuffled SFT phase, we move to the RL stage shown in Fig.~\ref{fig:framework}(d) and adopt Group Relative Policy Optimization (GRPO). For every question--answer pair $(q,a)$ the old policy $\pi_{\theta_{\text{old}}}$ samples a group of $G$ responses $\{o_i\}_{i=1}^{G}$. The reward for each response is defined as
\begin{equation}
\mathcal{R} = \mathcal{R}_{\text{format}} + \mathcal{R}_{\text{classification}},
\end{equation}
where $\mathcal{R}_{\text{classification}} = 1$ if the predicted class matches the ground-truth label and $0$ otherwise.  
Given the corresponding rewards $\{\mathcal{R}_i\}_{i=1}^{G}$, GRPO computes the relative advantage
\begin{equation}
\hat{A}_{i,t} = \frac{\mathcal{R}_i - \mathrm{mean}(\{\mathcal{R}_i\})}{\mathrm{std}(\{\mathcal{R}_i\})}.
\end{equation}

The optimization objective is
\begin{equation}
\begin{aligned}
\mathcal{J}_{\text{GRPO}}(\theta) &= \mathbb{E}_{(q,a)\sim\mathcal{D},\{o_i^G\}\sim\pi_{\theta_{\text{old}}}} \Bigg[\frac{1}{G}\sum_{i=1}^{G}\frac{1}{|o_i|}\sum_{t=1}^{|o_i|} \\
&\min\!\big(r_{i,t}(\theta)\hat{A}_{i,t},\,\mathrm{clip}(r_{i,t}(\theta),1-\varepsilon,1+\varepsilon)\hat{A}_{i,t}\big)\Bigg],
\end{aligned}
\end{equation}
with
\begin{equation}
r_{i,t}(\theta)=\frac{\pi_\theta(o_{i,t}\mid q,o_{i,<t})}{\pi_{\theta_{\text{old}}}(o_{i,t}\mid q,o_{i,<t})}.
\end{equation}

While the original GRPO adds a KL term $D_{\text{KL}}(\pi_\theta\|\pi_{\text{ref}})$ to curb large policy updates, we drop it under data-scarce conditions to avoid suppressing exploration.

\vspace{-0.6em}
\section{Experiments}

We conduct our evaluation and training using four widely adopted multimodal FAS datasets: WMCA \cite{wmca}, SURF \cite{surf}, CeFA \cite{cefa}, and PADISI \cite{padisi}. To assess the domain generalization capability of our model under multimodal settings, we follow the cross-domain evaluation protocol proposed in MMDG \cite{MMDG2024}, which includes several sub-protocols covering scenarios such as fixed modality, missing modality, and limited source domains. Detailed configurations are provided in supplementary material.

\vspace{-0.5em}
\subsection{Implementation Details}

We adopt Qwen2.5VL \cite{qwen2_5_2025} as our base multimodal large language model. All supervised fine-tuning and reinforcement learning stages are trained for a fixed 500 steps with a constant learning rate of 1e-6. For fair comparison, we categorize competing approaches into four groups: (1) Uni-modal DG methods that extend the input to multi-modal, (2) Multi-modal FAS methods, (3) Multi-modal DG FAS methods, and our proposed (4) interpretable Multi-modal DG FAS methods. This classification enables a more structured and meaningful evaluation for the FAS task.
% across different dimensions of

\vspace{-0.5em}
\subsection{Cross-Dataset Testing}
\noindent\textbf{Complete Modality Scenario.} \quad  
Protocol 1 is designed to evaluate model performance across unseen domains using multimodal data from varied scenarios. For example, the sub-protocol \textbf{CPS $\rightarrow$ W} represents that we take \textbf{C}, \textbf{P}, and \textbf{S} as training sets, while \textbf{W} is testing set. 
% 表\ref{table:P1}显示，在可解释多模态域泛化人脸防欺骗任务中，零样本基准模型 Qwen2.5VL-3B 的平均 HTER 高达 33.46%；经过 SFT 微调后，该指标降至 23.25%，但引入 RL 阶段后却反弹至 34.37%，暴露出 SFT+RL 范式在数据受限场景下的失效风险。相比之下，我们提出的 PA-FAS 将平均 HTER 进一步压至 15.73%，刷新同类方法最佳纪录，证明在仅约 800 条高质量结构化推理路径的驱动下，PA-FAS 能够超越依赖约 3.5 万条缺乏有效标注的原始数据所获得的泛化性能，显著缓解了数据稀缺与标注噪声带来的双重挑战。
As shown in Table~\ref{table:P1}, on the interpretable domain generalization multimodal FAS task, the zero-shot baseline Qwen2.5VL-3B yields an average HTER of 33.46\%. After SFT, this figure falls to 23.25\%, yet it rebounds sharply to 34.37\% once the RL stage is added, revealing the risk of SFT+RL collapse under data-scarce conditions. By contrast, our PA-FAS drives the average HTER down to 15.21\%, setting a new best-in-class record. This demonstrates that with only $\approx$800 high-quality structured reasoning paths, PA-FAS surpasses the generalization performance achieved by $\approx$35,000 raw data lacking reliable annotations, effectively mitigating the dual challenges of data scarcity and label noise.

\begin{table}[t]
\vspace{-0.9em}
% ========== 左侧宽表格 ==========
\centering
\caption{\small{Cross-dataset testing under limited source domain scenarios (Protocol 3) among CeFA-CeFA (C), PADISI USC (P), CASIA-SURF (S), and WMCA (W). }}
\vspace{-0.9em}
\resizebox{0.49\textwidth}{!}{
\begin{tabular}{l|cc|cc|cc}
    \toprule[1.3pt]
    \multirow{2}{*}{\textbf{Method}}  & \multicolumn{2}{c|}{\textbf{CW$\rightarrow$PS}}&\multicolumn{2}{c|}{\textbf{PS$\rightarrow$CW}}&\multicolumn{2}{c}{\textbf{Average}} \\
    \cmidrule(r){2-3} \cmidrule(r){4-5} \cmidrule(r){6-7}
    & HTER(\%)$\downarrow$ & AUC(\%)$\uparrow$ & HTER(\%)$\downarrow$ & AUC(\%)$\uparrow$& HTER(\%)$\downarrow$ & AUC(\%)$\uparrow$\\
    \midrule
    \multicolumn{7}{c}{\textbf{Uni-modal DG (Concat + 1*1 Conv)}}\\
    \midrule
    SSDG \textcolor{gray}{[CVPR'20]}&25.34&80.17&46.98&54.29&35.66&67.23\\
    SSAN \textcolor{gray}{[CVPR'22]}&26.55&80.06&39.10&67.19&32.82&73.62\\
    SA-FAS \textcolor{gray}{[CVPR'23]}&25.20&81.06&36.59&70.03&61.79&30.89\\
    IADG \textcolor{gray}{[CVPR'23]}&22.82&83.85&39.70&63.46&31.26&73.65\\
    FLIP \textcolor{gray}{[IJCAI'22]}&15.92&92.38&\underline{23.85}&\underline{83.46}&19.88&87.92\\
    \midrule
    \multicolumn{7}{c}{\textbf{Multi-modal FAS}}\\
    \midrule
    ViT \textcolor{gray}{[ICLR'20]}&42.66&57.80&42.75&60.41&42.70&59.10\\
    AMA \textcolor{gray}{[IJCV'24]}&29.25&76.89&38.06&67.64&33.65&72.26\\
    VP-FAS \textcolor{gray}{[TDSC'24]}&25.90&81.79&44.37&60.83&35.13&71.31\\
    ViTAF \textcolor{gray}{[ECCV'22]}&29.64&77.36&39.93&61.31&34.78&69.33\\
    MM-CDCN \textcolor{gray}{[CVPR'20]}&29.28&76.88&47.00&51.94&38.14&64.41\\
    CMFL \textcolor{gray}{[CVPR'21]}&31.86&72.75&39.43&63.17&35.64&67.96\\
    CLIP \textcolor{gray}{[ICML'21]}&19.36&90.57&29.98&79.22&24.67&84.89\\
    \midrule
    \multicolumn{7}{c}{\textbf{Multi-modal DG FAS}}\\
    \midrule
    MMDG \textcolor{gray}{[CVPR'24]}&20.12&88.24&36.60&70.35&28.36&79.3\\
    % MMDG++ \cite{mmdgplus2025}&10.67&95.95&21.55&86.73\\
    DADM \textcolor{gray}{[ICCV'25]}&12.61&93.81&\textbf{20.40}&\textbf{89.51}&\underline{16.50}&\textbf{91.66}\\
    % MMDA \cite{mmda2025}&7.52&96.84&6.30&98.35\\
    \midrule
    \multicolumn{7}{c}{\textbf{Interpretable Multi-modal DG FAS}}\\
    \midrule
    Qwen2.5-VL-3B \textcolor{gray}{[ARXIV'25]}&34.26&68.92&45.52&51.34&39.80&60.13\\
    Qwen2.5-VL-3B-SFT&\underline{0.60}&\underline{99.94}&33.76&69.94&17.18&84.94\\
    Qwen2.5-VL-3B-SFT+GRPO&\textbf{0.60}&\textbf{99.94}&33.78&69.71&17.19&84.82\\
    \textbf{PA-FAS (Ours)}&2.53&99.54&28.75&77.13&\textbf{15.64}&\underline{88.33}\\
    \bottomrule[1.3pt]
\end{tabular}}
\vspace{-0.8em}
\label{table:P3}
\end{table}

\begin{table}[t]
% ========== 左侧宽表格 ==========
\centering
%\vspace{-0.6em}
\caption{\small{Ablation on augmentation under the SFT+RL paradigm. }}
\vspace{-0.9em}
\resizebox{0.44\textwidth}{!}{
\begin{tabular}{l|c|c}
    \toprule[1.3pt]
    \textbf{Data} & HTER(\%)$\downarrow$ & AUC(\%)$\uparrow$\\
    \midrule
    Low-level Augmentation Data&34.37&68.68\\
    High-level Augmentation Data&44.49&52.95\\
    Reasoning Path Augmentation Data&\textbf{24.45}&\textbf{83.17}\\
    \bottomrule[1.3pt]
\end{tabular}}
% 不同数据集增强方法在SFT+RL训练范式下的性能对比
\vspace{-1.3em}
\label{table:Ablation PA}
\end{table}

\vspace{0.2em}
\noindent\textbf{Missing Modality Scenario During Testing.} \quad   
In Protocol 2, for each LOO sub-protocol of Protocol 1, we design three test-time missing-modal scenarios to validate the model’s performance when modalities are missing. 
% 表\ref{table:P2}表明，在多种模态缺失的设定下，无论模型处于零样本还是原始数据训练后的状态，其在各缺失场景中的性能曲线几乎重合，暴露出缺乏高质量思维链标注时，模型均仅依赖 RGB 图像、未能真正利用多模态信息的缺陷。而我们提出的方法缓解了这一局面，并将平均 HTER 从 33.46% 压至 15.85%。值得注意的是，当仅保留 RGB 单模态时，性能反而略优于 DEPTH 或红外单模态缺失的场景，暗示其他模态在当前训练条件下可能对 RGB 造成一定干扰。
Table~\ref{table:P2} shows that under various modality-missing conditions the performance curves of Qwen2.5VL-3B in its zero-shot and raw-data-trained states almost overlap, revealing that without high-quality chain-of-thought annotations the model relies solely on RGB images and fails to leverage multi-modal cues. Our proposed approach alleviates this limitation, driving the average HTER down from 33.46\% to 15.85\%. Interestingly, retaining only the RGB modality even yields a slight improvement over cases where DEPTH or infrared is individually missing, suggesting that the other modalities currently introduce a mild interference to the RGB signal.

% 表\ref{table:P3}显示，在两种限制源域的测试子协议下，我们方法相较零样本基线的 HTER 分别骤降 31.73\% 与 16.77\%，充分验证即使在源域数据受限的严苛场景中，依旧能够稳定输出优异性能。尤其是在有限标注数据的条件下，我们的路径增强方法依然能够在限制源域的情境下，取得与大量未标注数据相当的优异性能，这也充分展现了我们方法在数据利用效率上的优势。
\vspace{0.2em}
\noindent\textbf{Limited Source Domain Scenario.} \quad   
In Protocol 3, we limit the number of source domains by proposing two subprotocols, namely $CW \rightarrow PS$ and $PS \rightarrow CW$. As shown in Table~\ref{table:P3}, our method slashes HTER by 31.73\% and 16.77\% compared with the zero-shot baseline, demonstrating robust and superior performance even when source-domain data are severely limited. Especially under scenarios of limited labeled data, our path augmentation method still achieves performance comparable to that obtained with large amounts of unlabeled data, even in the context of restricted source domains, indicating our advantage of data efficiency.

\begin{table}[t]
% ========== 左侧宽表格 ==========
\vspace{-0.9em}
\caption{\small{Ablation on Path-Augmented data under shuffling. }}
\vspace{-0.9em}
\label{table:Ablation shuffle}
\centering
\resizebox{0.32\textwidth}{!}{
\begin{tabular}{l|c|c}
    \toprule[1.3pt]
    \textbf{Method} & HTER(\%)$\downarrow$ & AUC(\%)$\uparrow$\\
    \midrule
    w/o Shuffle&24.45&83.17\\
    w/ Shuffle Path&24.12&84.23\\
    w/ Shuffle Answer&\textbf{15.21}&\textbf{89.13}\\
    \bottomrule[1.3pt]
\end{tabular}}
% 路径增强数据在三种打乱方法下的性能对比
\vspace{-1.3em}
\end{table}

\begin{figure}[t]
%\vspace{-0.6em}
\centering
\includegraphics[width=0.48\textwidth]{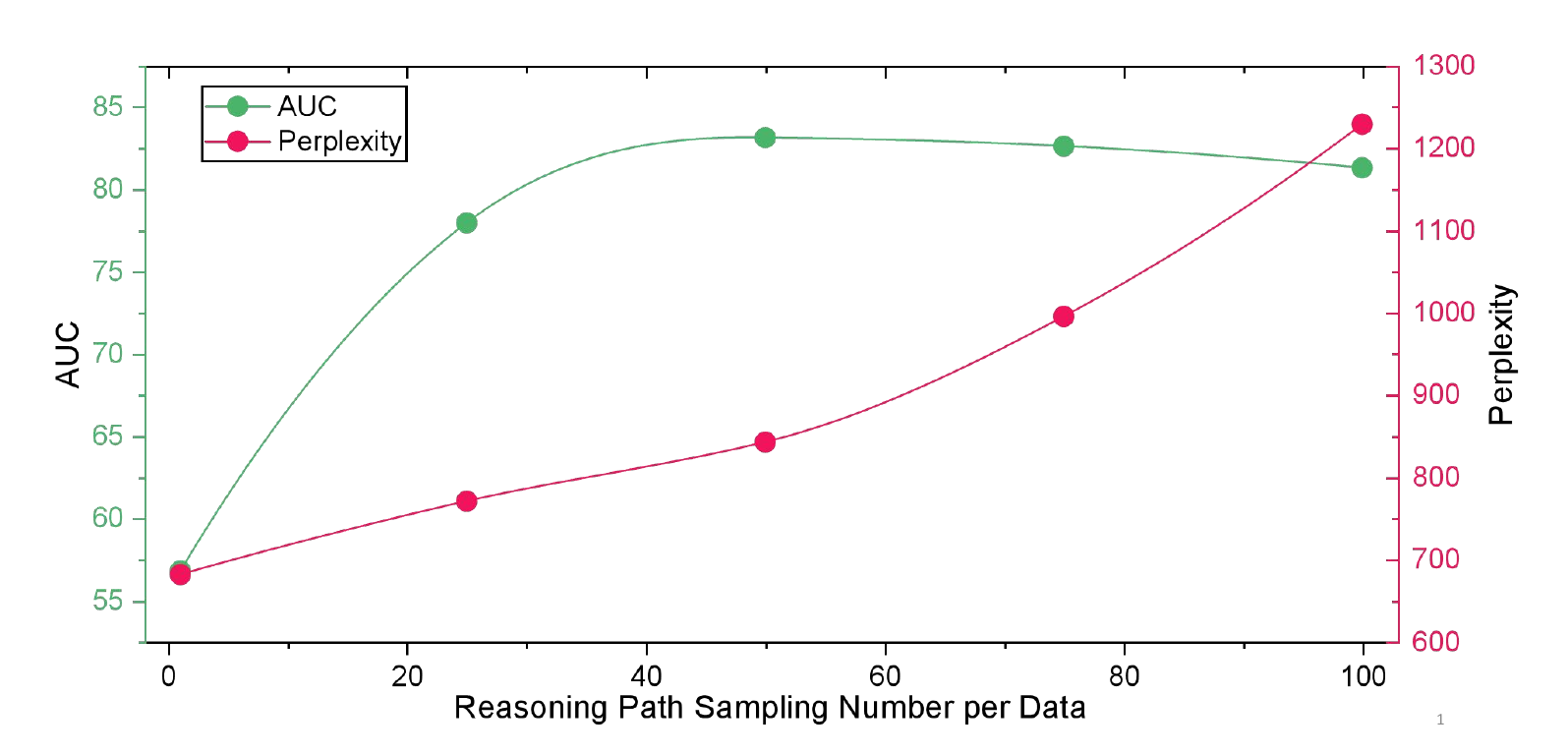}
\vspace{-2.3em}
% 不同推理路径采样数量下的AUC和Perplexity示意图。随着每个数据的采样路径数量不断增加，模型的困惑度（Perplexity）迅速上升。受此影响，AUC值先上升到一定水平，随后开始缓慢下降。
\caption{\small{A diagram of AUC and Perplexity vs. reasoning path sampling number $N$. As sampling paths increase, perplexity rises sharply, causing AUC to peak and then decline slowly.}}
\label{fig:ablation}
\vspace{-1.1em}
\end{figure}

\vspace{-0.5em}
\subsection{Ablation Study}

\noindent\textbf{Effectiveness of Reasoning Path Augmentation.} \quad  
% 表\ref{table:Ablation PA}显示，仅约 800 条经推理路径增强的样本在 SFT+RL 训练后，将 HTER 进一步拉低 9.92%，显著优于使用约 3.5 万条原始数据的模型；与 token 级增强的对比结果进一步表明，这一提升源于有效推理路径的扩张，而非简单的多样化扰动。
As shown in Table~\ref{table:Ablation PA}, training on approximately 800 reasoning-path-augmented data under the SFT+RL paradigm reduces HTER by 9.92\% compared with using roughly 35,000 raw data, demonstrating a substantial edge in both data efficiency and performance. A further comparison with token-level augmentation confirms that this gain stems from the expansion of valid reasoning paths rather than superficial diversity.

\vspace{0.2em}
\noindent\textbf{Effectiveness of Answer Shuffling Mechanism.} \quad  
% 表\ref{table:Ablation shuffle}表明，仅依靠推理路径增强虽扩大了有效数据，却仍无法避免推理捷径：路径随机打乱策略的 HTER 甚至略高于不扰动基线，暴露出捷径隐患。引入答案随机打乱后，HTER 骤降 9.24\%，充分验证该机制能够切断捷径，显著提升模型学习效果。
As shown in Table~\ref{table:Ablation shuffle}, although reasoning-path augmentation expands the pool of valid data, it still falls victim to reasoning shortcuts: path-level shuffling even yields a slightly higher HTER than the unshuffled baseline, highlighting the persistence of such shortcuts. In contrast, introducing answer shuffling decreases HTER by 9.24\%, conclusively demonstrating its ability to sever these shortcuts and markedly improve model learning.

\vspace{0.2em}
\noindent\textbf{Impact of Sampling Numbers per Data.} \quad
% 我们通过实验研究了推理路径增强方法在不同采样数量下的性能表现，结果如图 \ref{fig:ablation} 所示。实验发现，随着采样数量的增加，模型的困惑度急剧上升，表明过多的推理路径会阻碍模型的收敛。当采样数量 $N\approx50$ 时，模型的准确率达到峰值，这表明此时模型对单个数据的推理路径学习达到了最优状态。采样数量低于 50 时，准确率随推理路径增加而显著提升；超过 50 时，困惑度增加导致准确率逐渐下降。这一结果表明，采样数量对模型性能有显著影响。适中的采样数量能使模型充分利用推理路径信息，从而达到较高的准确率；而过多的采样则会导致路径信息冗余，使模型陷入困惑，进而降低性能。因此，合理选择采样数量对于优化模型性能至关重要。
We conduct studies on reasoning path enhancement under different sampling number $N$, as shown in Fig. \ref{fig:ablation}. The results indicate that as the sampling quantity increases, the model's perplexity rises sharply, suggesting that an excessive number of reasoning paths hinders model convergence. When $N\approx 50$, the model achieves the best performance. However, when it exceeds 50, the increase in perplexity leads to a gradual decline in AUC, indicating that an appropriate sampling quantity enables the model to fully utilize reasoning path information to achieve higher accuracy, while excessive samples result in redundant path information, causing model confusion and limited performance. 
%Therefore, it is crucial to choose an appropriate sampling quantity to optimize model performance.
%When the sampling quantity is below 50, the AUC increases significantly with the addition of reasoning paths. 

\vspace{-1.0em}
\section{Conclusion}
\vspace{-0.1em}
% 本工作通过引入推理路径增强方法与答案随机打乱机制，在有限标注数据条件下实现了 SFT + RL 训练范式在多模态面部防伪任务中的有效应用，初步验证了该方法在可解释性、多模态融合与跨域泛化三方面协同建模的可行性。该策略为构建统一、可信的 FAS 训练框架提供了新的路径选择。 尽管取得了一定进展，当前方法仍存在若干局限性：一方面，相较于各自领域内的专用模型，在可解释性、多模态建模或泛化性能上尚未达到最优表现；另一方面，由于标注数据受限，训练过程在稳定性与效率上仍有待提升，同时在多模态数据的充分利用上仍有所进步空间。未来的研究将聚焦于奖励函数的精细设计、强化学习策略的优化，以及低成本高效的数据标注方案的探索，以进一步提升模型性能与应用可行性。
%Nevertheless, compared with specialized models in each dimension, our method still falls short of state-of-the-art interpretability, multimodal integration, or generalization performance. Moreover
In this paper, we introduce a reasoning-path augmentation strategy together with an answer-shuffling mechanism, enabling the SFT+RL paradigm to be effectively applied to multimodal FAS under scarce annotations. It provides initial evidence that interpretability, multimodal fusion, and cross-domain generalization can be jointly modeled in a unified and trustworthy FAS training framework. However, the limited labeled data renders the training process less stable and efficient, and the utilization of multimodal information remains sub-optimal. Future works focus on refining the reward function design, optimizing RL strategies, and exploring low-cost, high-efficiency data-annotation schemes to further enhance performance and practical deployability.

\bibliography{aaai2026}

\end{document}